\documentclass[11pt,a4paper]{article}
\usepackage{times,latexsym}
\usepackage{url}
\usepackage[T1]{fontenc}
\usepackage[hidelinks]{hyperref}
\usepackage{graphicx}
\usepackage{subfigure}
\usepackage{tabularx}
\usepackage{arydshln}
\usepackage{amssymb,amsfonts}

\usepackage{amsmath}
\usepackage{multirow}
\usepackage{enumitem}
\usepackage{color}

\usepackage[acceptedWithA]{tacl2018v2}

\usepackage{xspace,mfirstuc,tabulary}

\def\eqsref#1#2{Eqs.~\ref{eq:#1}--\ref{eq:#2}}

\long\def\eat#1{\ignorespaces}

\def\figref#1{Figure~\ref{fig:#1}}

\def\tabref#1{Table~\ref{tab:#1}}

\def\secref#1{Section~\ref{sec:#1}}
\def\seclabel#1{\label{sec:#1}\label{p:#1}}
\def\eqref#1{Eq.~\ref{eqn:#1}}

\def\dnrm#1{\mbox{$_{\hbox{\scriptsize #1}}$}}

\long\def\eat#1{\ignorespaces}

\newif\iftaclinstructions
\taclinstructionsfalse 
\iftaclinstructions
\renewcommand{\confidential}{}
\renewcommand{\anonsubtext}{(No author info supplied here, for consistency with
TACL-submission anonymization requirements)}
\newcommand{\instr}
\fi

\iftaclpubformat 

\else

\fi

\title{Attentive Convolution: \\Equipping CNNs with RNN-style Attention Mechanisms}

\author{Wenpeng Yin \\
University of Pennsylvania\\
{\tt wenpeng@seas.upenn.edu}\\\And
Hinrich Sch\"utze\\
Center for Information and Language Processing\\
LMU Munich, Germany\\}

\date{}
\newcommand{\modelname}{\textsc{AttConv}}
\newcommand{\scitail}{\textsc{SciTail}}
\newcounter{notecounter}
\newcommand{\enotesoff}{\long\gdef\enote##1##2{}}
\newcommand{\enoteson}{\long\gdef\enote##1##2{{
\stepcounter{notecounter}
\large\bf
\hspace{1cm}\arabic{notecounter} $<<<$ ##1: ##2
$>>>$\hspace{1cm}}}}
\enoteson
\enotesoff

\begin{document}

\maketitle
\begin{abstract}
In NLP, convolutional neural networks (CNNs)
have benefited less  than recurrent neural
networks (RNNs) from attention mechanisms. We hypothesize
that this is because the attention in
CNNs  has been mainly implemented as
\emph{attentive pooling} (i.e., it is applied to pooling) rather than as
\emph{attentive convolution} (i.e., it is integrated into convolution).
Convolution is the differentiator of CNNs in that it
can powerfully model the higher-level
representation of a word by taking into account its
\emph{local fixed-size} context in the input text $t^x$. In this work, we
propose an attentive convolution network,
\modelname. It  extends the
  context scope of the convolution operation, deriving
higher-level features for a word not only from
local context, but also
 information extracted from nonlocal context by the
 attention mechanism commonly used in RNNs.
This nonlocal context
can come (i)
from parts of the input text $t^x$ that are distant
or
(ii) from \emph{extra} (i.e., external) contexts $t^y$.
Experiments on  sentence
 modeling with zero-context (sentiment analysis), single-context (textual entailment) and multiple-context (claim verification) demonstrate the effectiveness of
\modelname\enspace in sentence representation learning with
the incorporation of context. In particular, attentive
convolution outperforms attentive pooling and is a strong competitor to popular attentive RNNs.\footnote{\url{https://github.com/yinwenpeng/Attentive\_Convolution}}
\end{abstract}

\enote{hs}{
\begin{tabular}{l|lll}
& RNNs & AttentivePool & AttentivConv\\\hline
  source & $t^x$ & $t^x$ & $t^x$\\
  focus & $t^a$ &$t^a $ & $t^a$\\
weightee & $t^a$ &$t^x $ & $t^a$\\
  beneficiary & $t^x$ & $t^x$&$t^x$
  \end{tabular}

In this paper, we
conceptualize the
attention mechanism as having four components: source,
focus, weightee and beneficiary. The source consists of one
or more ``query''
vectors, e.g., the current hidden state of the decoder in
neural machine translation (NMT).
The focus is the set of vectors that are queried (e.g., the
hidden states of the encoder in NMT) to compute attention weights.
The weightee is the set of hidden states (e.g., the hidden
states of the encoder in NMT) whose
attention-weighted sum we compute. The beneficiary is the
set of vectors we interpret in the context of this
attention-weighted sum. In NMT, the beneficiary is
the hidden state of the decoder.

Attention has been extremely successful in RNNs.
The main contribution of this paper is that we
are the first work that enables CNNs to acquire this
powerful attention mechanism.
There has been prior work on attention in CNNs,
attentive pooling, but attentive pooling uses
a weak version of attention in which the weightee is
$t^x$, not $t^a$. In other words, the attention-weighted sum
in attentive pooling is a representation of $t^x$, not a
representation of additional valuable context from outside
of $t^x$.
}

\section{Introduction}
Natural language processing (NLP) has benefited greatly from
the resurgence of deep neural networks (DNNs), due to their
high performance with less need of engineered features. A
DNN typically is composed of a stack of non-linear
transformation layers, each generating a hidden
representation for the input by projecting the output of a
preceding layer into a new space. To date, building a single
and static representation to express an input across
diverse problems is far from satisfactory.
Instead, it is preferable that the representation of the input vary
in different
application scenarios. In response, attention mechanisms
\cite{DBLPGraves13,DBLPGravesWD14} are proposed to
dynamically focus on parts of the input that are expected to
be more specific to the problem. They are mostly implemented
based on fine-grained alignments between two pieces of
objects, each emitting a dynamic soft-selection to the
components of the other, so that the selected elements
dominate in the output hidden
representation. Attention-based DNNs have demonstrated good
performance on many tasks.



Convolutional neural networks (CNNs,
\newcite{lecun1998gradient}) and recurrent neural networks
(RNNs, \newcite{journalsElman90}) are  two important
types of DNNs.
Most work on attention has been done for
RNNs. Attention-based RNNs typically take three types of
inputs to make a decision at the current step: (i) the current
input state, (ii) a
representation of local context (computed unidirectionally or
bidirectionally, \newcite{entail2016}) and
(iii) the
attention-weighted sum of hidden states corresponding to
nonlocal context (e.g., the hidden states of the encoder in
neural machine translation \cite{DBLPBahdanauCB14}).
An important
question, therefore, is whether  CNNs can benefit from such
an attention mechanism as well, and how. This is our
\emph{technical motivation}.

Our second motivation is natural language understanding
(NLU). In generic sentence modeling \emph{without extra context}
\cite{collobert2011natural,kalchbrenner2014convolutional,kim2014convolutional},
CNNs learn sentence representations by composing word
representations that are conditioned on a local context
  window. We believe that attentive convolution is needed for some
  NLU tasks that are essentially sentence modeling within
  contexts. Examples: textual entailment (\newcite{DRSZ13},
Is a hypothesis  true given a premise as
the single context?) and claim verification (\newcite{DBLPfever05355},
Is a claim  correct given extracted
evidence snippets from a text corpus as the context?). Consider the
\scitail\enspace \cite{scitail} textual entailment examples
in \tabref{scitailexample}; here the input text $t^x$ is the
hypothesis and each premise is a context text $t^y$.
And consider the illustration of claim verification in
\figref{claimverification}; here the input text $t^x$ is the
claim and $t^y$ can consist of multiple pieces of
context.  In both cases, we would like the representation
of $t^x$ to be context specific.

\begin{table}
 \setlength{\belowcaptionskip}{-15pt}
 \setlength{\abovecaptionskip}{5pt}
  \centering
  \footnotesize
  \begin{tabular}{l|c}
   \multicolumn{1}{c|}{ premise, modeled as context $t^y$ } & \\\hline
Plant cells have structures that animal cells lack. & 0\\
 Animal cells do not have cell walls. & 1\\
	The cell wall is not a freestanding structure.&	0\\
 Plant cells possess a cell wall, animals never. & 1
\end{tabular}
  \caption{Examples of
    four premises for the hypothesis
   $t^x$ = ``\emph{A cell wall is not present in animal cells.}'' in
    \textsc{SciTail}  dataset. Right column (hypothesis's label): ``1'' means \emph{true}, ``0'' otherwise.}\label{tab:scitailexample}
\end{table}


In this work, we propose attentive convolution networks,
\modelname, to  model a sentence (i.e., $t^x$)
either in
intra-context (where $t^y=t^x$) or extra-context (where $t^y\neq t^x$ and
$t^y$ can have many pieces) scenarios.  In the
\emph{intra-context} case (sentiment analysis, for example),
\modelname\enspace extends the local context window of
standard CNNs to cover the entire input text $t^x$.  In the
\emph{extra-context} case, \modelname\enspace extends the
local context window to cover accompanying contexts $t^y$.


For a
convolution operation over a window in $t^x$ such as (\emph{left}$\dnrm{context}$,
\emph{word}, \emph{right}$\dnrm{context}$), we first compare the
representation of \emph{word} with all hidden states in the context $t^y$
to get an attentive context representation
\emph{att}$\dnrm{context}$, then convolution filters derive a higher-level
representation for \emph{word}, denoted as
\emph{word}$_\mathrm{new}$,
by integrating \emph{word} with three pieces of
context: \emph{left}$\dnrm{context}$, \emph{right}$\dnrm{context}$ and
\emph{att}$\dnrm{context}$.
We interpret this
attentive convolution in two perspectives. (i) For intra-context, a
higher-level word representation \emph{word}$\dnrm{new}$ is learned
by considering the local (i.e., \emph{left}$\dnrm{context}$ and
\emph{right}$\dnrm{context}$) as well as nonlocal (i.e.,
\emph{att}$\dnrm{context}$) context. (ii) For extra-context,
  \emph{word}$\dnrm{new}$ is generated to
represent
 \emph{word}, together with its cross-text alignment \emph{att}$\dnrm{context}$, in
the context \emph{left}$\dnrm{context}$ and
\emph{right}$\dnrm{context}$. In other words, the decision
for the \emph{word} is made based on the connected hidden states of cross-text aligned terms, with local context.

We apply \modelname\enspace to three sentence modeling tasks with variable-size context: a large-scale
Yelp sentiment classification task \cite{DBLPLinFSYXZB17}
(intra-context, i.e., no additional context),
\scitail\enspace textual entailment \cite{scitail} (single extra-context) and claim verification \cite{DBLPfever05355}  (multiple extra-contexts). \modelname\enspace
outperforms competitive DNNs with and without attention
and gets state-of-the-art on the three tasks.

Overall, we make the following contributions:

\textbullet\enspace This is the first work that equips convolution filters with the attention mechanism commonly employed in RNNs.

\textbullet\enspace We distinguish and build flexible modules -- attention source, attention focus and attention beneficiary -- to greatly advance the expressivity of attention mechanisms in CNNs.

\textbullet\enspace \modelname\enspace provides a new way to broaden the originally constrained scope of filters in conventional CNNs. Broader and richer context comes from
  either external context (i.e., $t^y$)  or the sentence itself (i.e., $t^x$).

\textbullet\enspace \modelname\enspace shows its flexibility and effectiveness in sentence modeling with variable-size context.



\begin{figure}
 \setlength{\belowcaptionskip}{-15pt}
 \setlength{\abovecaptionskip}{5pt}
\centering
\includegraphics[width=0.48\textwidth]{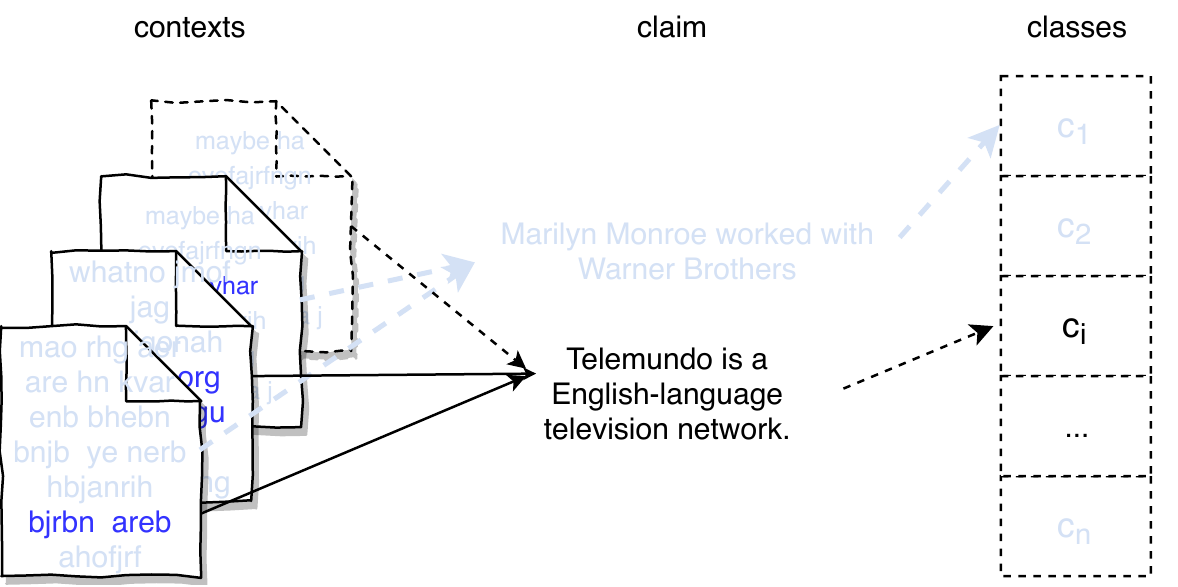}
\caption{Verify claims in contexts}\label{fig:claimverification}
\end{figure}

\section{Related Work}\label{sec:relatedwork}
In this section we discuss attention-related
DNNs  in NLP, the most relevant work for our paper.

\begin{figure}[!h]
\centering
\includegraphics[width=0.45\textwidth]{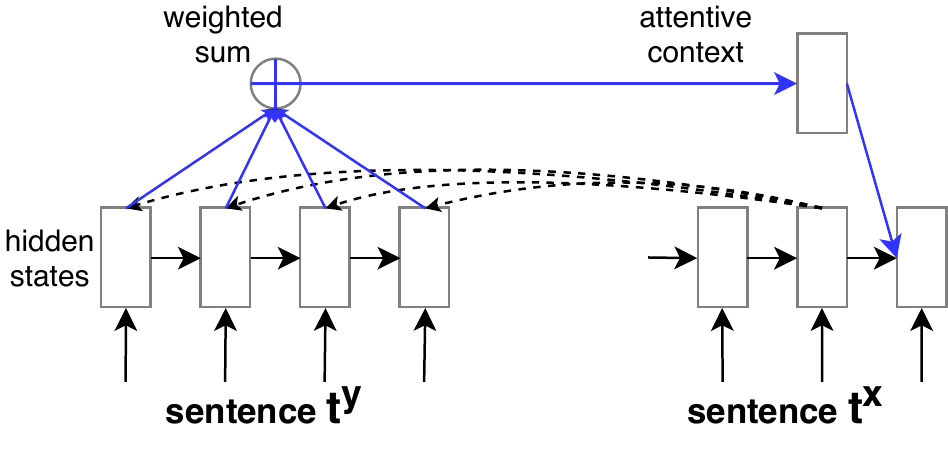}
\caption{A simplified illustration of attention mechanism in RNNs.}\label{fig:attentivernn}
\end{figure}

\begin{figure}[!h]
\centering
\includegraphics[width=0.48\textwidth]{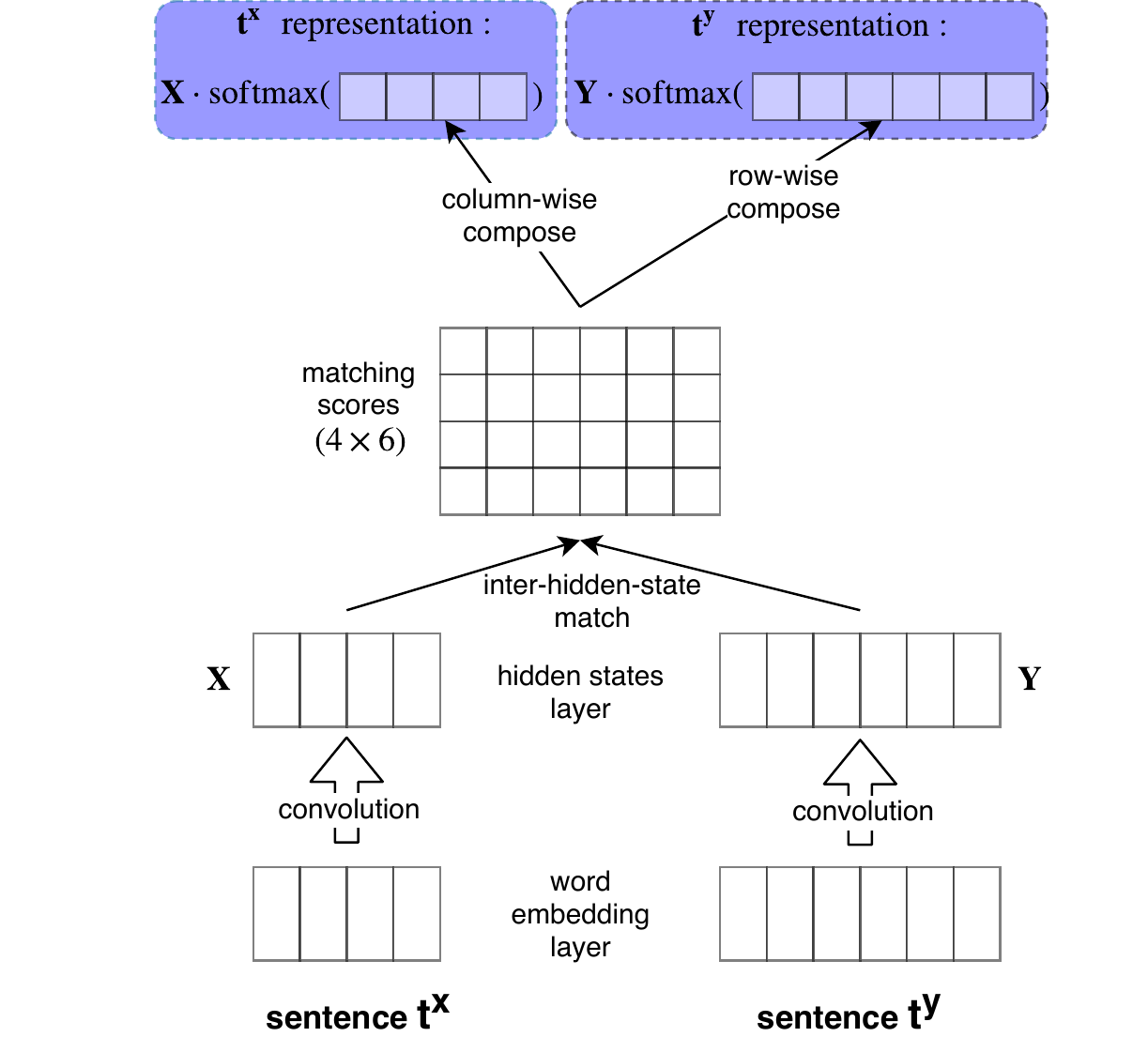}
\caption{Attentive pooling, summarized from ABCNN \protect\cite{DBLPYinSXZ16} and APCNN \protect\cite{santos2016attentive}.}\label{fig:attentivepooling}
\end{figure}

\subsection{RNNs with Attention}
\newcite{DBLPGraves13} and \newcite{DBLPGravesWD14} first
introduced a differentiable attention mechanism that allows
RNNs to focus on different parts of the
input.
This  idea has  been broadly explored in RNNs, shown in \figref{attentivernn}, to deal with \textbf{text generation}, such as neural machine translation \cite{DBLPBahdanauCB14,DBLPLuongPM15,DBLPLibovickyH17,DBLPKimDHR17}, response generation in social media \cite{DBLPShangLL15},
document reconstruction \cite{li2015hierarchical} and document
summarization \cite{DBLPNallapatiZSGX16}, \textbf{machine comprehension} \cite{DBLPHermannKGEKSB15,DBLPKumarIOIBGZPS16,DBLPXiongMS16,DBLPWangJ17a,DBLPXiongZS16,DBLPeoKFH16,rnet} and \textbf{sentence relation classification}, such as textual entailment \cite{entail2016,wang2015learning,DBLP0001DL16,DBLPWangHF17,DBLPChenZLWJI17} and answer sentence selection \cite{DBLPMiaoYB16}.

We try to explore the RNN-style attention mechanisms in CNNs, more specifically, in convolution.

\begin{figure*}
 \setlength{\belowcaptionskip}{-10pt}
 \setlength{\abovecaptionskip}{0pt}
\centering
\subfigure[Light attentive convolution layer] { \label{fig:lightacnn}
\includegraphics[width=0.37\textwidth]{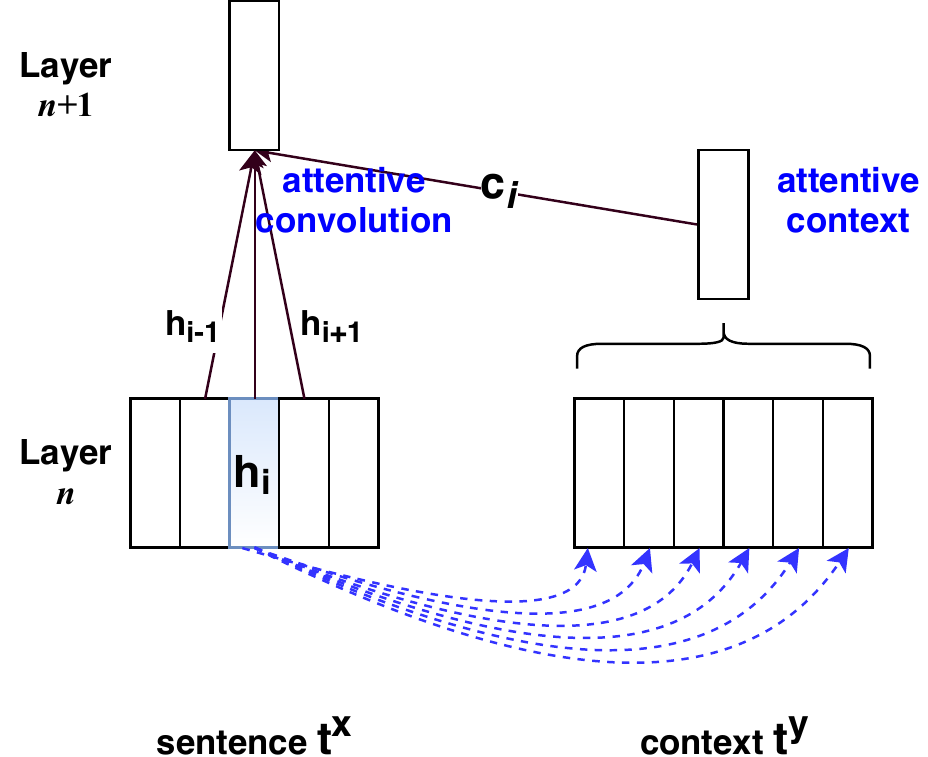}
}
\subfigure[Advanced attentive convolution layer] { \label{fig:advancedacnn}
\includegraphics[width=0.58\textwidth]{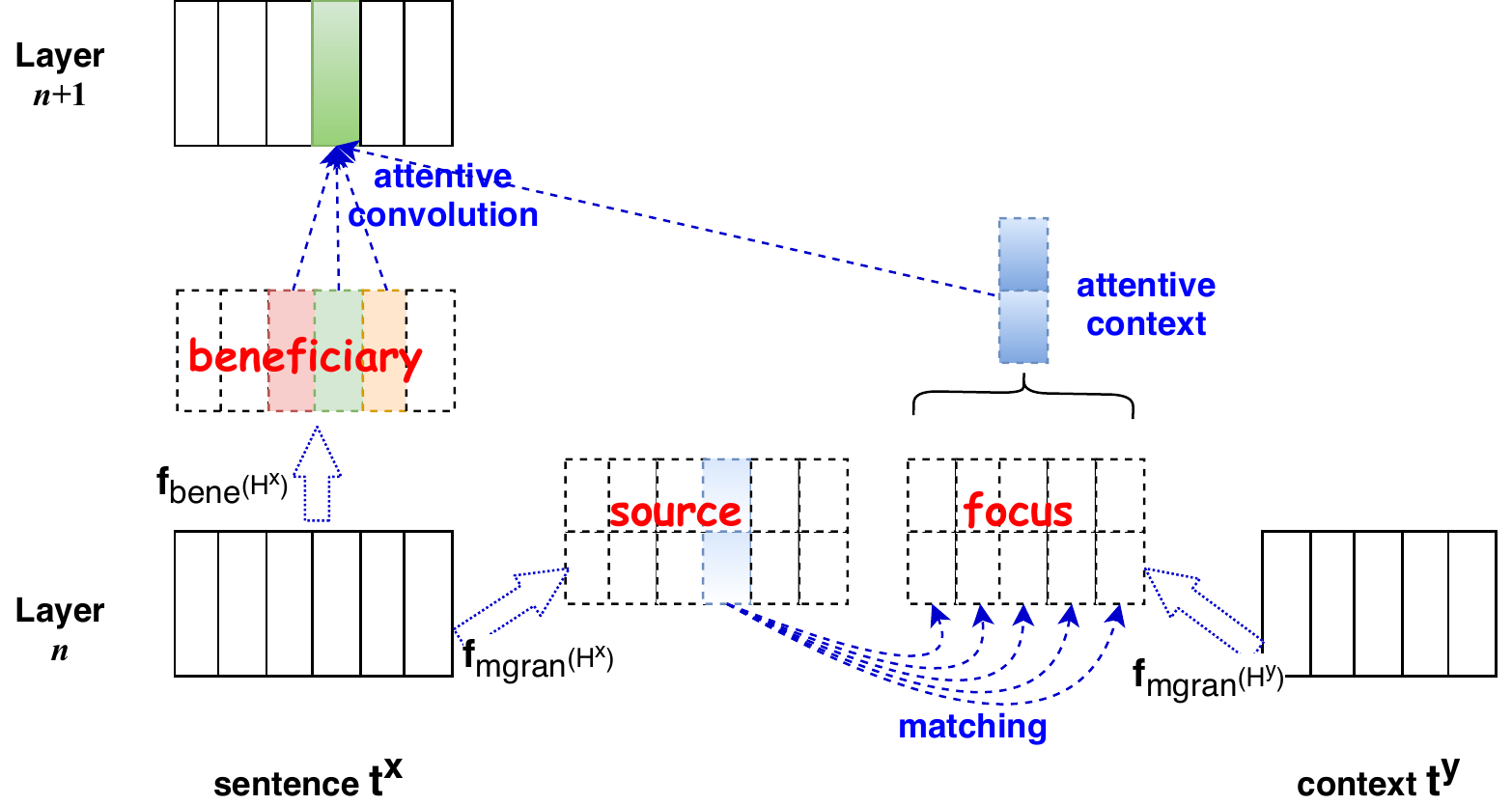}
}
\caption{\modelname\enspace models sentence $t^x$ with context $t^y$}
\label{fig:acnn}
\end{figure*}

\subsection{CNNs with Attention}

In NLP, there is little work on attention-based
CNNs. \newcite{DBLPGehringAGYD17} propose an attention-based
convolutional seq-to-seq model for machine translation. Both
the encoder and decoder are hierarchical convolution
layers. At the $n^{th}$  layer  of the decoder, the  output
hidden state of a convolution queries each of the encoder-side
hidden states, then a weighted sum of all encoder hidden
states is added to the decoder hidden state, finally this
updated hidden state is fed to the convolution  at layer
$n+1$.  Their attention implementation relies on the
existence of a multi-layer convolution structure --
otherwise the weighted context from the encoder side could
not play a role in the decoder. So essentially their
attention is achieved \emph{after} convolution. In contrast,
we aim to modify  the vanilla convolution, so that CNNs with attentive convolution can be applied more broadly.

We discuss two systems that are
representative
of CNNs that
implement the attention \emph{in pooling} (i.e., the convolution
is still not affected):
\cite{DBLPYinSXZ16,santos2016attentive}, illustrated in \figref{attentivepooling}.
 Specifically, these two systems work on two  input sentences, each with a set of hidden states generated by a convolution layer, then
each sentence will learn a weight for every hidden state  by comparing this hidden state with all hidden states in the other sentence,
finally each input sentence obtains a representation by a
weighted mean pooling over all its hidden states. The
  core component -- weighted mean pooling -- was referred to as ``attentive pooling'', aiming to yield the sentence representation.

In contrast to attentive convolution,
attentive pooling does not
connect directly the hidden states of cross-text aligned phrases
 in a fine-grained manner to the final decision making;  only the matching scores
contribute to the final weighting in mean pooling.
This important distinction
between attentive convolution and attentive pooling
is further discussed in
\secref{modelanalysis}.

Inspired by the attention mechanisms in RNNs, we
  assume that it is the \emph{hidden states of aligned phrases}
  rather than \emph{their matching scores} that can better
  contribute to  representation learning and decision
  making. Hence, our \emph{attentive convolution  differs from
  attentive pooling} in that it
  uses \emph{attended hidden states from extra context} (i.e., $t^y$) \emph{or broader-range
  context}  within $t^x$ \emph{to participate in the convolution}. In experiments, we
  will show its superiority.





\section{\modelname\enspace Model}\label{sec:model}
We use  bold uppercase, e.g.,
$\mathbf{H}$, for matrices;
bold lowercase,
e.g.,
$\mathbf{h}$,  for vectors;
bold lowercase with index,
e.g.,
$\mathbf{h}_i$,  for columns of
 $\mathbf{H}$; and non-bold lowercase for
scalars.



To start, we assume that a piece of
text $t$ ($t \in \{t^x, t^y\}$) is represented as a sequence of
hidden states $\textbf{h}_i\in\mathbb{R}^d$
($i=1,2,\ldots, |t|$), forming feature map
$\mathbf{H}\in\mathbb{R}^{d\times |t|}$, where $d$ is the
dimensionality of hidden states. Each hidden state $\textbf{h}_i$ has its left context $\mathbf{l}_{i}$ and
right context $\textbf{r}_{i}$. In  concrete CNN systems, context $\mathbf{l}_{i}$ and
$\textbf{r}_{i}$ can cover multiple adjacent hidden states;
we set $\mathbf{l}_{i}=\textbf{h}_{i-1}$ and
$\textbf{r}_{i}=\textbf{h}_{i+1}$ for simplicity in the following description.

We now describe light and advanced versions of
\modelname. Recall that \modelname\enspace aims  to compute a
representation for $t^x$ in a way that convolution filters   encode not only local
context, but also attentive context over $t^y$.

\subsection{Light \modelname}
Figure \ref{fig:lightacnn} shows the light
version of
\modelname. It differs in two key points -- (i) and
(ii) -- both from the basic convolution
layer that models a single piece of text and  from  the Siamese CNN that
models two text pieces in parallel. (i) A
matching  function
determines how relevant each hidden
state in the context $t^y$ is to the current
hidden state $\textbf{h}^x_i$ in sentence $t^x$. We then compute an
average of the hidden states in  the context $t^y$, weighted by the
matching scores,
to get the
attentive context $\textbf{c}^x_i$ for $\textbf{h}^x_i$. (ii)
The convolution for position $i$ in $t^x$ integrates hidden
state $\textbf{h}^x_i$ with \emph{three sources of context}: left
context $\textbf{h}^x_{i-1}$, right context $\textbf{h}^x_{i+1}$ and
attentive context $\textbf{c}^x_i$.

\paragraph{Attentive Context.} First, a
 function generates a matching score $e_{i,j}$
between a hidden state in $t^x$ and a hidden state in $t^y$
by (i)
dot product:
\begin{equation}\label{eq:docproduct}
\abovedisplayskip=2pt
\belowdisplayskip=2pt
e_{i,j} = (\textbf{h}^x_i)^T\cdot \textbf{h}^y_j
\end{equation}
or (ii) bilinear form:
\begin{equation}
\abovedisplayskip=2pt
\belowdisplayskip=2pt
e_{i,j} =  (\textbf{h}^x_i)^T\mathbf{W}_e \textbf{h}^y_j
\end{equation}
(where $\mathbf{W}_{e}\in\mathbb{R}^{d\times d}$) or (iii) additive projection:
\begin{equation}
\abovedisplayskip=4pt
\belowdisplayskip=4pt
e_{i,j} = (\mathbf{v}_e)^T\cdot\mathrm{tanh}(\mathbf{W}_e\cdot \mathbf{h}^x_i + \mathbf{U}_e \cdot \mathbf{h}^y_j)
\end{equation}
where $\mathbf{W}_{e},\mathbf{U}_{e}\in\mathbb{R}^{d\times d}$ and  $\mathbf{v}_{e}\in\mathbb{R}^{d}$.

Given the matching scores, the attentive context $\mathbf{c}^x_i$ for hidden state $\mathbf{h}^x_i$ is the weighted average of all hidden states in $t^y$:
\begin{equation}
\abovedisplayskip=2pt
\belowdisplayskip=2pt
\mathbf{c}^x_i = \sum_j \mathrm{softmax}(\mathbf{e}_{i})_j\cdot \mathbf{h}^y_j
\end{equation}
We refer to the concatenation of attentive contexts
$[\mathbf{c}^x_1; \ldots; \mathbf{c}^x_i; \ldots;
  \mathbf{c}^x_{|t^x|}]$ as the
feature map
$\mathbf{C}^x\in\mathbb{R}^{d\times|t^x|}$ for $t^x$.

\paragraph{Attentive Convolution.}
After attentive context has been computed,
a position $i$ in the sentence $t^x$  has a hidden state $\mathbf{h}^{x}_i$,   the left
context $\mathbf{h}^{x}_{i-1}$, the right context
$\mathbf{h}^{x}_{i+1}$ and the attentive context
$\mathbf{c}^{x}_i$. Attentive convolution then  generates the higher-level hidden state at position $i$:
{\small
\begin{align}\label{eq:aconv}
\abovedisplayskip=0pt
\belowdisplayskip=0pt
\mathbf{h}^{x}_{i,\mathrm{new}}=\mathrm{tanh}(&\mathbf{W}\cdot [\mathbf{h}^{x}_{i-1},\mathbf{h}^{x}_i,\mathbf{h}^{x}_{i+1},\mathbf{c}^{x}_i]+\mathbf{b}) \\
\label{eq:aconvsplit}
=\mathrm{tanh}(&\mathbf{W}^1\cdot [\mathbf{h}^{x}_{i-1},\mathbf{h}^{x}_i,\mathbf{h}^{x}_{i+1}]+ \nonumber\\
&\mathbf{W}^2\cdot \mathbf{c}^{x}_i+\mathbf{b})
\end{align}
}where $\mathbf{W}\in\mathbb{R}^{d\times 4d}$ is the concatenation of $\mathbf{W}^1\in\mathbb{R}^{d\times 3d}$ and $\mathbf{W}^2\in\mathbb{R}^{d\times d}$, $\mathbf{b}\in\mathbb{R}^d$.

As Equation \ref{eq:aconvsplit} shows, Equation
\ref{eq:aconv} can be achieved by summing up the results of
two separate and parallel convolution steps before the
non-linearity. The first is still a
standard convolution-without-attention
over
feature map $\mathbf{H}^{x}$ by filter width 3 over the window
($\mathbf{h}^{x}_{i-1}$, $\mathbf{h}^{x}_i$,
$\mathbf{h}^{x}_{i+1}$). The second
is a convolution  on the feature
map $\mathbf{C}^{x}$, i.e., the attentive context, with
filter width 1, i.e., over each
$\mathbf{c}^{x}_i$; then sum up the results
element-wise and add a bias term and the non-linearity. This
divide-then-compose strategy makes the attentive convolution easy to
implement in practice with no need to create a new feature
map, as required in Equation
\ref{eq:aconv}, to integrate $\mathbf{H}^{x}$ and $\mathbf{C}^{x}$.

It is worth mentioning that
$\mathbf{W}^1\in\mathbb{R}^{d\times 3d}$ corresponds to the
filter parameters of a vanilla CNN and the only added
parameter here is $\mathbf{W}^2\in\mathbb{R}^{d\times
  d}$, which only depends  on the hidden size.

This light \modelname\enspace shows the basic principles of
using RNN-style attention mechanisms in convolution. Our
experiments show that this light version of \modelname --
even though it incurs a limited increase of parameters (i.e.,
$\mathbf{W}^2$) --  works much better than the vanilla
Siamese CNN and some of the pioneering attentive RNNs.  The
following two considerations show that there is space to
improve its expressivity.

(i) \emph{Higher-level or more abstract representations are
  required in subsequent layers.}  We find that directly
forwarding the hidden states in $t^x$ or $t^y$ to the
matching process does not work well in some tasks. Pre-learning
some more high-level or abstract representations helps in
subsequent learning phases.

\begin{table}
 \setlength{\belowcaptionskip}{-15pt}
 \setlength{\abovecaptionskip}{3pt}
\small
\setlength{\tabcolsep}{1pt}
  \centering
  \begin{tabular}{l |l}
  \multicolumn{1}{c|}{role} & \multicolumn{1}{c}{text}\\\hline
premise &   Three firefighters \textbf{come out} of subway station\\\hdashline
\multirow{2}{*}{hypothesis}&  Three firefighters \textbf{putting out} a fire \textbf{inside}\\
& of a subway station \\
  \end{tabular}
\caption{Multi-granular alignments required in TE}\label{tab:snlimgran}
\end{table}

(ii) \emph{Multi-granular alignments are preferred in the
  interaction modeling between $t^x$ and $t^y$.}  Table
\ref{tab:snlimgran} shows another example of textual
entailment. On the unigram level, ``out'' in the premise matches
with ``out'' in the hypothesis perfectly, while ``out'' in
the premise is contradictory to ``inside'' in the hypothesis. But their context snippets -- ``come out'' in the premise and
``putting out a fire'' in the hypothesis -- clearly indicate
they are not semantically equivalent. And the gold
conclusion for this pair is ``neutral'', i.e., the
hypothesis is possibly true. Therefore, matching should be
conducted across phrase granularities.

We now  present  advanced \modelname. It is more
expressive and modular, based on the two foregoing
considerations (i) and (ii).

\subsection{Advanced \modelname}\label{sec:advancedmodel}

\newcite{heike2017}  distinguish between \textbf{focus} and
\textbf{source} of attention. The \emph{focus} of attention
is the layer of the network that is reweighted by attention
weights. The \emph{source} of attention is the information
source that is used to compute the attention weights.
\newcite{heike2017}
showed that increasing the scope of the
attention source is beneficial. It possesses some  preliminary principles of the query/key/value distinction by \newcite{DBLPVaswaniSPUJGKP17}.  Here we further extend this principle to
define \textbf{beneficiary} of attention -- the feature map
(labeled ``beneficiary''
in  Figure \ref{fig:advancedacnn})
that is contextualized by \emph{the attentive
  context} (labeled ``attentive context''
in  Figure \ref{fig:advancedacnn}).
In light attentive
convolutional layer (Figure \ref{fig:lightacnn}), the
source of attention is hidden states in sentence $t^x$, the
focus of attention is hidden states of the context $t^y$, the
beneficiary of attention is again the hidden states of $t^x$, i.e., it is
identical to the source of
attention.

We now try to distinguish these three
concepts further to promote the expressivity of an attentive
convolutional layer. We call it ``advanced
\modelname'', see
Figure \ref{fig:advancedacnn}.
It differs from the light version in
three ways: (i) \emph{attention source} is learned by function
$f_{\mathrm{mgran}}(\mathbf{H}^x)$, feature map
$\mathbf{H}^x$ of $t^x$ acting as input; (ii)
\emph{attention focus} is learned by function
$f_{\mathrm{mgran}}(\mathbf{H}^y)$, feature map
$\mathbf{H}^y$ of context $t^y$ acting as input; (iii)
\emph{attention beneficiary} is learned by function
$f_{\mathrm{bene}}(\mathbf{H}^x)$, $\mathbf{H}^x$ acting as
input.  Both functions $f_{\mathrm{mgran}}()$ and
$f_{\mathrm{bene}}()$ are based on a \emph{gated
  convolutional function} $f_{\mathrm{gconv}}()$:
\begin{align}
\overline{\mathbf{o}}_i &= \mathrm{tanh}(\mathbf{W}_h\cdot \mathbf{i}_i + \mathbf{b}_h)\\
\mathbf{g}_i &= \mathrm{sigmoid}(\mathbf{W}_g\cdot \mathbf{i}_i + \mathbf{b}_g)\\
f_{\mathrm{gconv}}(\mathbf{i}_i)  &= \mathbf{g}_i \cdot \mathbf{u}_i + (1-\mathbf{g}_i)\cdot \overline{\mathbf{o}}_i
\end{align}
where $\mathbf{i}_i$ is a composed  representation, denoting  a generally defined input phrase $[\cdots, \mathbf{u}_i, \cdots]$ of arbitrary length with $\mathbf{u}_i$ as the central unigram-level hidden state, the gate $\mathbf{g}_i$ sets a trade-off between the unigram-level input $\mathbf{u}_i$ and the temporary output $\mathbf{\overline{o}}_i$ at the phrase-level.
We elaborate these modules in the remainder of this subsection.

\textbf{Attention Source.} First, we present a general instance of generating source of attention by function
$f_{\mathrm{mgran}}(\mathbf{H})$, learning word representations in multi-granular context. In our system, we consider granularities one and three, corresponding to uni-gram hidden state and tri-gram hidden state. For the uni-hidden state case, it is  a gated convolution layer:
\begin{equation}
\abovedisplayskip=2pt
\belowdisplayskip=2pt
\mathbf{h}^x_{\mathrm{uni}, i} = f_{\mathrm{gconv}}(\mathbf{h}^x_i)
\end{equation}
For tri-hidden state case:
\begin{equation}
\abovedisplayskip=3pt
\belowdisplayskip=3pt
\mathbf{h}^x_{\mathrm{tri},i} = f_{\mathrm{gconv}}([\mathbf{h}^x_{i-1},\mathbf{h}^x_i, \mathbf{h}^x_{i+1}])
\end{equation}
Finally, the overall hidden state at position $i$ is the concatenation of $\mathbf{h}_{\mathrm{uni}, i}$ and $\mathbf{h}_{\mathrm{tri}, i}$:
\begin{equation}
\abovedisplayskip=3pt
\belowdisplayskip=3pt
\mathbf{h}^x_{\mathrm{mgran},i} = [\mathbf{h}^x_{\mathrm{uni}, i},\mathbf{h}^x_{\mathrm{tri}, i}]
\end{equation}
i.e., $f_{\mathrm{mgran}}(\mathbf{H}^x)=\mathbf{H}^x_{\mathrm{mgran}}$.

Such kind of comprehensive hidden state can encode
the semantics of multigranular spans at a position, such as
``out'' and ``come out of''. Gating here implicitly
  enables cross-granular alignments in subsequent attention
  mechanism as it sets highway connections
  \cite{DBLPSrivastavaGS15} between the input granularity
  and the output granularity.

\textbf{Attention Focus.}
For simplicity,
we use the same architecture for the \emph{attention
  source} (just introduced) and  for the \emph{attention focus},
$t^y$;
i.e., for the attention focus:
$f_{\mathrm{mgran}}(\mathbf{H}^y)=\mathbf{H}^y_{\mathrm{mgran}}$.
See
Figure \ref{fig:advancedacnn}.
Thus, the focus of
attention will participate in the matching process as well
as be reweighted to form an attentive context vector. We leave
exploring different architectures
for attention source and focus for future work.

Another benefit of multi-granular hidden states in attention
focus is to keep structure information in the context vector. In
standard attention mechanisms in RNNs, all hidden states are
average-weighted as a context vector, the order information
is missing. By introducing hidden states
of larger granularity
into
CNNs that keep the
local order or structures, we boost the attentive effect.

\textbf{Attention Beneficiary.}
In our system, we simply use $f_{\mathrm{gconv}}()$ over uni-granularity   to learn a more abstract representation over the current hidden representations in $\mathbf{H}^x$, so that
\begin{equation}
\abovedisplayskip=3pt
\belowdisplayskip=3pt
f_{\mathrm{bene}}(\mathbf{h}^x_i) = f_{\mathrm{gconv}}(\mathbf{h}^x_i)
\end{equation}
Subsequently, the \emph{attentive context vector}
$\mathbf{c}^x_i$ is generated based on  \emph{attention
  source} feature map $f_{\mathrm{mgran}}(\mathbf{H}^x)$ and
\emph{attention focus} feature map
$f_{\mathrm{mgran}}(\mathbf{H}^y)$, according to the
description of the light \modelname. Then
attentive convolution is conducted over \emph{attention
  beneficiary} feature map $f_{\mathrm{bene}}(\mathbf{H}^x)$
and the attentive context vectors $\mathbf{C}^x$ to get
a higher-layer feature map  for the sentence
$t^x$.

\subsection{Analysis}\label{sec:modelanalysis}


\textbf{Compared to the standard attention mechanism in RNNs,}
\modelname\enspace has a similar matching
function and a similar process of computing context
vectors, but differs in three
ways. (i) The discrimination of attention source, focus and
beneficiary improves expressivity. (ii) In
CNNs, the surrounding hidden states for a
concrete position are available, so the attention matching is
able to encode the left context as well as the right
context. In RNNs however, it needs bidirectional RNNs to
yield both left and right context representations. (iii) As
attentive convolution  can be implemented by
summing up two separate convolution steps
(\eqsref{aconv}{aconvsplit}), this
architecture provides both attentive representations and
representations computed without the use of attention. This strategy is helpful
in practice to use richer representations for some NLP problems. In contrast,
such
a clean modular  separation of representations computed with and without
attention is harder to realize in
attention-based RNNs.


\textbf{Prior attention mechanisms explored in CNNs} mostly
involve attentive pooling
\cite{DBLPYinSXZ16,santos2016attentive}, i.e., the
weights of the post-convolution pooling layer are
determined by attention. These weights come
from the matching process between hidden states of two text
pieces. However, a weight value is not informative enough to
tell the relationships between aligned terms. Consider a
textual entailment sentence pair for which we need to
determine whether ``inside $\longrightarrow$ outside''
holds.
The matching degree (take cosine similarity as example)
of these two words is high, e.g.,
$\approx$ .7 in Word2Vec \cite{DBLPMikolovSCCD13} and GloVe
\cite{pennington2014glove}. On the other hand, the matching score
between ``inside'' and ``in'' is lower: .31 in Word2Vec,
.46 in GloVe. Apparently, the higher number .7 does not mean that
``outside'' is more likely than ``in'' to be entailed by
``inside''. Instead, joint representations for aligned
phrases [$\mathbf{h}\dnrm{inside}$, $\mathbf{h}\dnrm{outside}$],
[$\mathbf{h}\dnrm{inside}$, $\mathbf{h}\dnrm{in}$] are more
informative and enable finer-grained reasoning than a
mechanism that can only transmit information downstream by
matching scores. We modify the conventional CNN filters so that ``inside'' can make the entailment decision by looking at  the representation of the counterpart term (``outside'' or ``in'') rather than a matching score.

A more damaging property of attentive pooling is the following. Even
if matching scores could  convey the
phrase-level entailment degree to some extent, matching weights, in fact, are
not leveraged to make the entailment decision directly;
instead, they are used to weight the sum of the output hidden
states of  a convolution as the global sentence
representation. In other words, fine-grained
entailment degrees are likely to be lost in the summation of
many vectors.  This illustrates why
attentive context vectors participating in the convolution
operation are expected to be more effective than
post-convolution attentive pooling (more explanations in
Section \ref{sec:entailtask}, see paragraph
``\textbf{Visualization}'').

\textbf{Intra-context attention \& extra-context attention.}
Figures \ref{fig:lightacnn}-\ref{fig:advancedacnn}
depict the modeling of a sentence $t^x$ with its context
$t^y$. This is a common application of attention mechanism
in literature; we call it extra-context attention. But
\modelname\enspace can also be applied to model
a single text input, i.e., intra-context attention.  Consider a
sentiment analysis example: ``\emph{With the 2017 NBA All-Star game in the books
I think we can all agree that this was definitely
one to remember. Not because of the three-point
shootout, the dunk contest, or the game itself but
because of the ludicrous trade that occurred after
the festivities}''; this example  contains
informative points at different locations (``remember'' and ``ludicrous''); conventional
CNNs' ability
to model  nonlocal dependency is limited due to
fixed-size filter widths. In  \modelname, we can
set $t^y = t^x$. The attentive context vector
then accumulates all related parts together for a
given position. In other words, our intra-context attentive
convolution is able to connect all related spans together to
form a comprehensive decision. This is a new way to
  broaden the scope of conventional filter widths: a filter
  now covers not only the local window, but also those spans
  that are related, but are  beyond the scope of the window.

\textbf{Comparison to Transformer.}\footnote{Our
  ``source-focus-beneficiary'' mechanism was inspired by
  \newcite{heike2017}.
  \newcite{DBLPVaswaniSPUJGKP17} later published
  the   Transformer model, which has a similar
``query-key-value'' mechanism.}
The ``focus'' in \modelname\enspace corresponds to
``key'' and ``value'' in Transformer; i.e., our versions of
``key'' and ``value'' are the same, coming from the context
sentence.  The ``query'' in Transformer corresponds to the
``source'' and ``beneficiary'' of \modelname; i.e.,
our model has two perspectives to utilize the context: one
acts as a real query (i.e., ``source'') to attend the
context, the other (i.e., ``beneficiary'') takes the
attentive context back to improve the learned representation
of itself.  If we reduce \modelname\ to  unigram
convolutional filters, it is pretty much a single
Transformer layer (if we neglect the positional encoding in Transformer and
unify the ``query-key-value'' and
``source-focus-beneficiary'' mechanisms).

\begin{table}
  \centering
  \begin{tabular}{c |l l| l}
     & \multicolumn{2}{c|}{systems}& \multicolumn{1}{c}{acc}\\ \hline \hline
    \multirow{4}{*}{\rotatebox{90}{\scriptsize{\begin{tabular}{c}w/o\\ attention\end{tabular}}}}
    &\multicolumn{2}{l|}{Paragraph Vector} & 58.43 \\
    &\multicolumn{2}{l|}{\small{Lin et al.\ Bi-LSTM}} & 61.99\\
    &\multicolumn{2}{l|}{\small{Lin et al.\ CNN}} & 62.05\\
    &\multicolumn{2}{l|}{\emph{MultichannelCNN (Kim)}}& \emph{64.62}\\\hline
    \multirow{5}{*}{\rotatebox{90}{\scriptsize{\begin{tabular}{c}with\\ attention\end{tabular}}}}
    & \multicolumn{2}{l|}{CNN+internal attention} & 61.43\\
    & \multicolumn{2}{l|}{ABCNN} & 61.36\\
    & \multicolumn{2}{l|}{APCNN} & 61.98\\
    & \multicolumn{2}{l|}{Attentive-LSTM} & 63.11\\
    &\multicolumn{2}{l|}{Lin et al.\ RNN Self-Att.} & 64.21\\\hline
    \multirow{2}{*}{\rotatebox{90}{\scriptsize{\begin{tabular}{c}\textsc{Att}\\ \textsc{Conv}\end{tabular}}}}
    &\multicolumn{2}{l|}{light} & 66.75\\
    & \multicolumn{2}{l|}{\enspace\enspace w/o convolution} & 61.34\\
    & \multicolumn{2}{l|}{advanced}& \textbf{67.36}$^*$

\end{tabular}
  \caption{System comparison of sentiment analysis on Yelp.
Significant
improvements over state of the art are marked
with $*$ (test of equal proportions, p $<$ .05). }\label{tab:yelp}
\end{table}
\section{Experiments}\label{sec:experiment}
We evaluate \modelname\ on sentence modeling in three scenarios: (i) Zero-context, i.e., intra-context; the same input sentence acts as $t^x$ as well as $t^y$; (ii) Single-context. Textual entailment -- hypothesis modeling with a single premise as the extra-context; (iii) Multiple-context. Claim verification -- claim modeling with multiple extra-contexts.

\subsection{Common setup and common baselines}
\seclabel{common}
All experiments share
a \textbf{common setup}.
The input is represented using 300-dimensional publicly available Word2Vec \cite{DBLPMikolovSCCD13} embeddings;
OOVs are randomly
initialized.
The architecture consists of the following four layers in sequence:
embedding, attentive
convolution, max-pooling and
logistic regression. The context-aware representation of $t^x$ is forwarded to the logistic regression layer.
We use AdaGrad
\cite{duchi2011adaptive} for training.
Embeddings are fine-tuned during
training. Hyperparameter values include: learning rate 0.01, hidden size 300, batch size 50, filter width 3.


All experiments are designed to explore comparisons in three aspects: (i) within \modelname, ``light'' vs. ``advanced'';
(ii) ``attentive convolution'' vs. ``attentive pooling''/``attention only'';
 (iii) ``attentive convolution'' vs. ``attentive RNN''.

To this end, we always  report ``light'' and ``advanced''
\modelname\ performance and compare against
five types of \textbf{common baselines}: (i) w/o context,
(ii) w/o-attention, (iii) w/o-convolution: Similar with the Transformer's principle \cite{DBLPVaswaniSPUJGKP17}, we discard the convolution operation in Equation \ref{eq:aconv} and forward the addition of the attentive context $\mathbf{c}_i^x$ and the $\mathbf{h}_i^x$  into a fully connected layer. To keep enough parameters, we stack totally four layers so that ``w/o-convolution'' has the same size of parameters as light-\modelname, (iv) with-attention: RNNs with attention and
CNNs with attentive pooling
and
(v) prior state of the art, \emph{typeset in italics}.

\subsection{Sentence modeling with zero-context: sentiment analysis}

We evaluate sentiment analysis on a
Yelp benchmark released by
\newcite{DBLPLinFSYXZB17}:  review-star pairs
in sizes 500K (train), 2000 (dev) and 2000 (test).
Most text
instances in this dataset  are long: 25\%, 50\%, 75\%
percentiles are  46, 81, 125 words, respectively.
The task is five-way classification: one to five stars.
The measure is accuracy. We use this benchmark because the
predominance of
long texts lets us  evaluate the system performance of encoding long-range context, and the system by \newcite{DBLPLinFSYXZB17} is directly related to \modelname\enspace in intra-context scenario.


\textbf{Baselines.} (i)
w/o-attention.
Three baselines from
\newcite{DBLPLinFSYXZB17}:
Paragraph Vector \cite{le2014distributed} (unsupervised sentence representation
learning), BiLSTM and CNN. We also
reimplement
MultichannelCNN
\cite{kim2014convolutional}, recognized as a
simple, but surprisingly strong sentence modeler.
(ii) with-attention. A vanilla ``Attentive-LSTM'' by \newcite{entail2016}.
``RNN Self-Attention''
\cite{DBLPLinFSYXZB17} is directly comparable to \modelname:
it  also uses intra-context attention.
 ``CNN+internal attention''
\cite{heike2017},
an intra-context attention idea similar to, but less
complicated than
\cite{DBLPLinFSYXZB17}. ABCNN/APCNN -- CNNs with attentive pooling.

\begin{figure}
\label{fig:yelppred}
 \setlength{\belowcaptionskip}{-5pt}
 \setlength{\abovecaptionskip}{2pt}
\includegraphics[width=0.9\columnwidth]{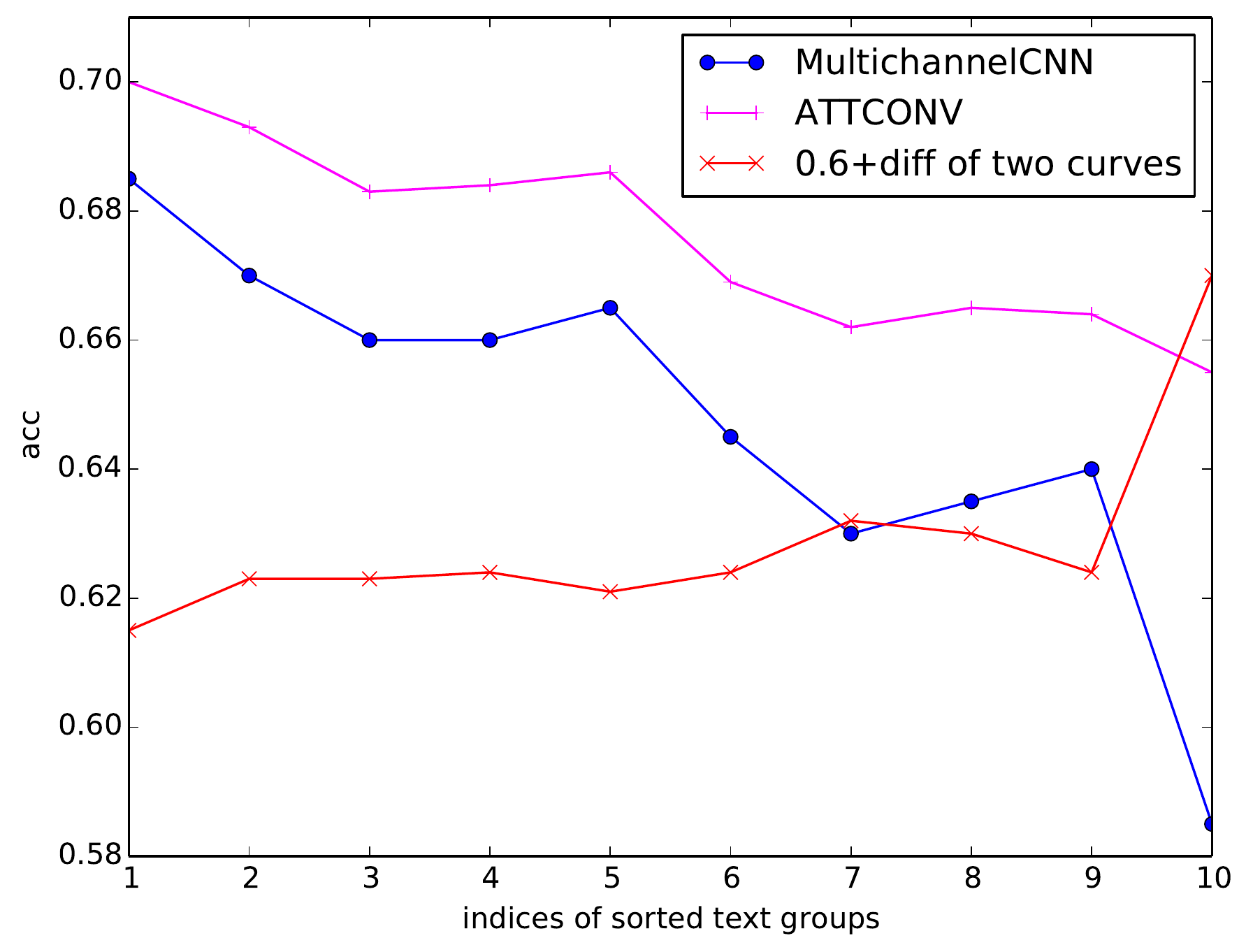}
\centering{}
\caption{\modelname\ vs.\ MultichannelCNN for groups of Yelp text with ascending text lengths.  \modelname\enspace performs more  robust across different lengths of text.}
\end{figure}

\textbf{Results and Analysis.}
Table \ref{tab:yelp} shows that advanced-\modelname\enspace surpasses its ``light'' counterpart, and gets significant improvement over the state of the art.

In addition, \modelname\enspace surpasses attentive pooling (ABCNN\&APCNN) with a big margin ($>$5\%) and outperforms the representative attentive-LSTM ($>$4\%).

Furthermore, it outperforms the two
self-attentive models: CNN+internal
attention \cite{heike2017} and RNN Self-Attention
\cite{DBLPLinFSYXZB17}, which are specifically designed for single-sentence modeling. \newcite{heike2017} generate an
attention weight for each CNN hidden state by a linear
transformation of the same hidden state, then compute
weighted average over all hidden states as the text
representation. \newcite{DBLPLinFSYXZB17} extend that idea
by generating \emph{a group of} attention
weight vectors, then RNN hidden states are averaged by those
diverse weighted vectors, allowing extracting different
aspects of the text into multiple vector representations.
Both works are essentially weighted mean pooling, similar to the
attentive pooling in
\cite{DBLPYinSXZ16,santos2016attentive}.

Next, we compare \modelname\enspace with MultichannelCNN,
the strongest baseline system (``w/o attention''), for different
length ranges to check if \modelname\enspace can really encode long-range context effectively. We  sort the
2000 test instances by length, then split
them into 10 groups, each consisting of 200
instances.
Figure \ref{fig:yelppred}
shows performance of \modelname\
vs.\ MultichannnelCNN.

We observe that \modelname\enspace consistently outperforms
MultichannelCNN for all
lengths. Furthermore, the improvement
over MultichannelCNN  generally increases with length.
This is evidence that
\modelname\enspace is more effectively  modeling long
text. This is likely due to  \modelname's capability
to
encode broader context in its filters.

\subsection{Sentence modeling with  a single context: textual entailment}\label{sec:entailtask}
\begin{table}
 \setlength{\belowcaptionskip}{-5pt}
 \setlength{\abovecaptionskip}{5pt}
  \centering
  \begin{tabular}{l|rrr}
   &  \multicolumn{1}{c}{\#instances}  & \#entail  & \#neutral\\\hline
	train & 23,596 & 8,602 & 14,994\\
    dev & 1,304 & 657 & 647\\
    test & 2,126 & 842 & 1,284\\\hline
    total & 27,026 & 10,101 & 16,925
\end{tabular}
\caption{Statistics of \scitail\enspace dataset}\label{tab:scitaildata}
\end{table}

\begin{table}
 \setlength{\belowcaptionskip}{-5pt}
 \setlength{\abovecaptionskip}{5pt}
  \centering
  \begin{tabular}{c|l|c}
  & \multicolumn{1}{c|}{systems} & acc\\\hline
  \multirow{5}{*}{\rotatebox{90}{\begin{tabular}{c}w/o\\ attention\end{tabular}}} &Majority Class &	60.4\\
  & w/o Context & 65.1\\
  &Bi-LSTM & 69.5\\
  &NGram model  &	70.6\\
  &Bi-CNN & 74.4\\\hdashline
  \multirow{6}{*}{\rotatebox{90}{\begin{tabular}{c} with\\ attention\end{tabular}}}
 & Enhanced LSTM 	&70.6\\
&Attentive-LSTM &71.5 \\
 & Decomp-Att  &72.3\\
 & DGEM &77.3\\\cdashline{2-3}
 	&APCNN &75.2\\
   & ABCNN &75.8\\\hdashline
 \multicolumn{2}{l|}{ \modelname-light}  &78.1\\
  \multicolumn{2}{l|}{ \enspace\enspace\enspace w/o convolution}  &75.1\\
 \multicolumn{2}{l|}{ \modelname-advanced}  &\textbf{79.2}
\end{tabular}
\caption{\modelname\enspace vs. baselines on \scitail\enspace dataset}\label{tab:scitailresults}
\end{table}

\begin{table*}
  \centering
   \small
  \begin{tabular}{c|l|c|l}
   \# & \multicolumn{1}{c|}{(Premise $t^y$, Hypothesis $t^x$) Pair} & G/P & Challenge \\\hline

  \multirow{3}{*}{1} & ($t^y$) These insects have 4 life stages, the egg, larva, pupa and adult. & \multirow{3}{*}{1/0} & \multirow{3}{1.5cm}{language conventions}\\
& ($t^x$) The sequence egg $-$$>$ larva $-$$>$ pupa $-$$>$ adult shows the  life cycle& &\\
&  of some insects.& & \\\hline

    \multirow{2}{*}{2} & ($t^y$) \ldots\ the notochord forms the backbone (or vertebral column). & \multirow{2}{*}{1/0} & \multirow{2}{1.5cm}{language conventions}\\
  & ($t^x$) Backbone is another name for the vertebral column. & & \\\hline

        \multirow{3}{*}{3} & ($t^y$) Water lawns early in the morning \ldots\ prevent evaporation. & \multirow{3}{*}{1/0} & \multirow{3}{*}{beyond text}\\
  & ($t^x$) Watering plants and grass in the early morning is a way to conserve water  & &\\
  & because smaller amounts of water evaporate in the cool morning.& & \\\hline

        \multirow{2}{*}{4} & ($t^y$) \ldots\ the SI unit \ldots\ for force is the Newton (N) and is defined as (kg$\cdot$m/s$^{-2}$ ). & \multirow{2}{*}{0/1} & \multirow{2}{*}{beyond text}\\
  & ($t^x$) Newton (N) is the SI unit for weight. & &  \\\hline

      \multirow{3}{*}{5} & \multirow{2}{11.5cm}{($t^y$) Heterotrophs get energy and carbon from living plants or animals (consumers) or from dead organic matter (decomposers).} & \multirow{3}{*}{0/1} & \multirow{3}{1.5cm}{discourse relation}\\
  &  & & \\
  & ($t^x$) Mushrooms get their energy from decomposing dead organisms. & &\\\hline

      \multirow{3}{*}{6} & ($t^y$) \ldots\ are a diverse  assemblage of three phyla of nonvascular plants, with  & \multirow{3}{*}{1/0} & \multirow{3}{1.5cm}{discourse relation}\\
  & about 16,000 species, that includes the mosses, liverworts, and hornworts. & & \\
  & ($t^x$) Moss is best classified as a nonvascular plant. & &

\end{tabular}
\caption{Error cases of \modelname\enspace in \scitail. ``\ldots'': truncated text. ``G/P'': gold/predicted label.}\label{tab:scitalerrorexample}
\end{table*}
\textbf{Dataset.} \textsc{SciTail}  \cite{scitail} is a
textual entailment benchmark designed specifically for a
real-world task: multi-choice question answering.
 All hypotheses $t^x$ were
obtained by rephrasing  (question, correct answer) pairs into single sentences, and premises $t^y$ are relevant web sentences retrieved by an
 information retrieval (IR) method.  Then the task is to determine whether  a hypothesis is \emph{true or not}, given a  premise as context. All ($t^x$, $t^y$) pairs are
annotated via crowdsourcing. Accuracy is reported.  Table \ref{tab:scitailexample}
shows examples and Table \ref{tab:scitaildata} gives  statistics.
\emph{By this construction,  a substantial
  performance improvement on \textsc{Sci\-Tail}  is equivalent to
  a better QA performance} \cite{scitail}. The hypothesis $t^x$ is the target sentence, and the premise $t^y$ acts as its context.

\begin{table}[h]
\setlength{\tabcolsep}{3pt}
  \centering
  \small
  \begin{tabular}{c|l | c| c}
    &\multicolumn{1}{c|}{\textbf{Systems}}& \#para & acc\\ \hline \hline
     \multirow{6}{*}{\rotatebox{90}{\begin{tabular}{c}w/o\\ attention\end{tabular}}} & majority class &0& 34.3\\
     &  w/o context (i.e., hypothesis only)&270K& 68.7\\
          &Bi-LSTM \cite{bowman2015large} &220K& 77.6\\
     & Bi-CNN &270K& 80.3\\
     & Tree-CNN \cite{mouP16} &3.5M& 82.1\\
     & NES \cite{DBLPYuM17a} &6.3M& 84.8\\\hline
     \multirow{8}{*}{\rotatebox{90}{\begin{tabular}{c}with\\ attention\end{tabular}}}&
     Attentive-LSTM (Rockt\"{a}schel)
     & 250K&83.5 \\
   & Self-Attentive \cite{DBLPLinFSYXZB17}&95M& 84.4\\
   & Match-LSTM (Wang \& Jiang) &1.9M& 86.1\\

   &LSTMN \cite{DBLP0001DL16} &3.4M& 86.3\\
   & Decomp-Att (Parikh) & 580K&86.8\\
   & Enhanced LSTM {\tiny\cite{DBLPChenZLWJI17}} &7.7M& 88.6\\\cdashline{2-3}
   & ABCNN \cite{DBLPYinSXZ16} &834K& 83.7\\
   & APCNN \cite{santos2016attentive} &360K& 83.9\\\hline
   \multicolumn{2}{l|}{\modelname\enspace -- light} &360K& 86.3\\
   \multicolumn{2}{l|}{\enspace\enspace\enspace w/o convolution} &360K& 84.9\\
   \multicolumn{2}{l|}{\modelname\enspace -- advanced} &900K& 87.8\\\hline
\multicolumn{2}{l|}{State-of-the-art \cite{DBLP05365}} &8M& \textbf{88.7}

\end{tabular}
\caption{Performance comparison on SNLI test. Ensemble systems are not included.}\label{tab:snlicomparison}
\end{table}

\paragraph{Baselines.}  Apart from the common baselines (see
\secref{common}), we  include  systems covered by \newcite{scitail}:
(i) Ngram Overlap: An overlap baseline,
considering lexical granularity such as unigrams, one-skip bigrams and one-skip
trigrams. (ii) Decomposable Attention Model (Decomp-Att)
\cite{DBLParikhT0U16}: Explore attention mechanisms to
decompose the task into  subtasks to solve in
parallel. (iii) Enhanced LSTM \cite{DBLPChenZLWJI17}:
Enhance LSTM by taking into account syntax and semantics from parsing
information.  (iv) DGEM \cite{scitail}: A decomposed graph
entailment model, the current state-of-the-art.


\begin{figure*}[!h]
  \setlength{\belowcaptionskip}{-4pt}
  \setlength{\abovecaptionskip}{5pt}
\centering
\subfigure[Visualization for features generated by \modelname's filters on sentence $t^x$ and $t^y$. A max-pooling, over filter\_1,  locates the phrase (A, dog, jumping, $\mathbf{c}^x_{dog}$), and locates the phrase (a, Frisbee, in, $\mathbf{c}^x_{Frisbee}$) via filter\_2. ``$\mathbf{c}^x_{dog}$'' (resp. $\mathbf{c}^x_{Fris.}$) --  the attentive context of ``dog'' (resp. ``Frisbee'') in $t^x$ -- mainly comes from ``animal'' (resp. ``toy'' and ``playing'') in $t^y$.]
{ \label{fig:visualization}
\includegraphics[width=0.98\textwidth]{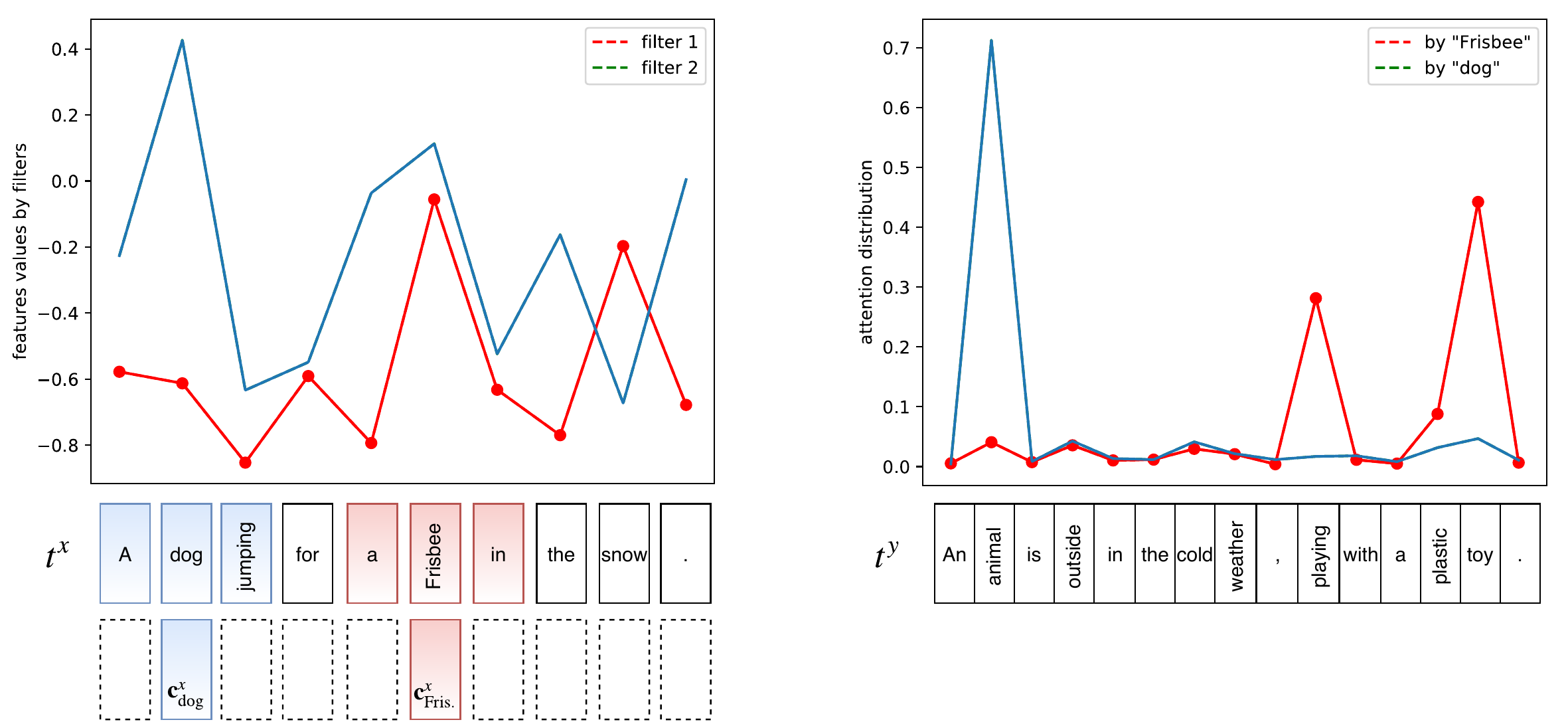}
}
\subfigure[Attention visualization for attentive pooling
  (ABCNN). Based on the words in $t^x$ and $t^y$, first, a
  convolution layer with filter width 3 outputs hidden
  states for each sentence, then each hidden state will
  obtain an attention weight for how well  this hidden state
  matches towards all the hidden states in the other
  sentence, finally all hidden states in each sentence will
  be weighted and summed up as the sentence
  representation. This visualization shows that the spans ``dog jumping for'' and ``in the snow'' in $t^x$ and the spans ``animal is outside'' and  ``in the cold'' in $t^y$ are most indicative to the entailment reasoning.]
{ \label{fig:visualizationpooling}
\includegraphics[width=0.98\textwidth]{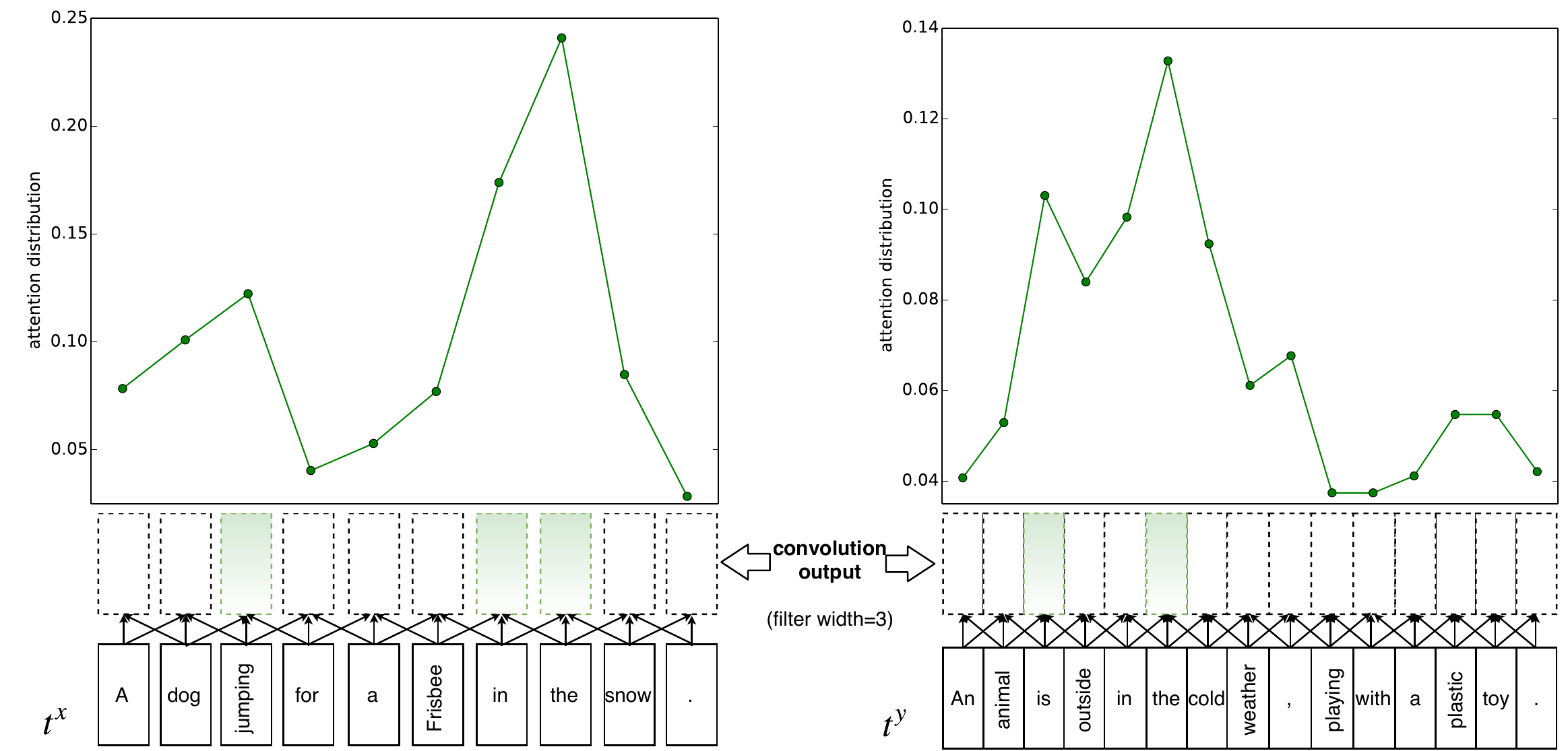}
}
\caption{Attention visualization for Attentive Convolution (top) and Attentive Pooling (bottom) between sentence $t^x$ = ``\emph{A dog jumping for a Frisbee in the snow.}'' (left) and sentence $t^y$ = ``\emph{An animal is outside in the cold weather, playing with a plastic toy.}'' (right).}
\label{fig:overallvisualization}
\end{figure*}


Table \ref{tab:scitailresults} presents \textbf{results} on \scitail.
(i) Within \modelname, ``advanced'' beats ``light'' by 1.1\%; (ii) ``w/o convolution'' and attentive pooling (i.e., ABCNN/APCNN) get lower performances by 3\%--4\%; (iii) More complicated attention mechanisms equipped into LSTM (e.g., ``attentive-LSTM'' and ``enhanced-LSTM'') perform even worse.


\textbf{Error Analysis.}  To better understand the \modelname\enspace in \scitail, we study some error cases listed in Table \ref{tab:scitalerrorexample}.

\emph{Language conventions.} Pair \#1 uses sequential commas (i.e., in ``the egg, larva, pupa and adult'')
or a special symbol sequence (i.e., in ``egg $-$$>$ larva $-$$>$ pupa $-$$>$ adult'') to form a set or sequence; pair \#2
has ``A (or B)'' to express the equivalence of A and
B. This challenge is expected to be handled by deep neural networks with specific training signals.

\emph{Knowledge beyond the text $t^y$.} In \#3,
``\emph{because smaller amounts of water evaporate in the
  cool morning}'' cannot be inferred from the premise $t^y$
directly. The main challenge in   \#4 is to
distinguish  ``weight'' from ``force'', which requires
background physical knowledge that is beyond the presented text
here and beyond the expressivity of word embeddings.

\emph{Complex discourse relation.} The premise in
\#5 has an ``or'' structure. In \#6, the inserted
phrase ``\emph{with about 16,000 species}'' makes the
connection between ``\emph{nonvascular plants}'' and
``\emph{the mosses, liverworts, and hornworts}'' hard to
detect. Both instances require the model to
decode the discourse relation.

\paragraph{\modelname\enspace on SNLI.} \tabref{snlicomparison} shows the comparison. We observe that: (i) classifying hypotheses without looking at premises, i.e., ``w/o context'' baseline, gets a big improvement over the ``majority baseline''. This verifies the strong bias in the hypothesis construction of SNLI dataset \cite{DBLP02324,DBLPPoliakNHRD18}. (ii) \modelname\enspace(advanced) surpasses all ``w/o attention'' baselines and  ``with attention'' CNN baselines (i.e., attentive pooling), obtaining a performance (87.8\%) that is close to the state of the art (88.7\%).

We also report the parameter size in SNLI as most baseline
systems did.  \tabref{snlicomparison} shows that, in
comparison to these baselines,
our
\modelname\enspace(light \& advanced) has a more limited number
of parameters, yet its performance is competitive.

\paragraph{Visualization.} In \figref{overallvisualization}, we visualize the attention mechanisms explored in attentive convolution (i.e., \figref{visualization}) and attentive pooling (i.e., \figref{visualizationpooling}).

\figref{visualization} explores the visualization of two kinds of features learned by light \modelname\enspace in SNLI dataset (most are short sentences with rich phrase-level reasoning): (i) $e_{i,j}$ in Equation \ref{eq:docproduct} (after softmax), which shows the \emph{attention distribution} over context $t^y$ by the hidden state $h^x_i$  in sentence $t^x$; (ii) $\mathbf{h}^x_{i,\mathrm{new}}$ in Equation \ref{eq:aconv} for $i=1,2,\cdots, |t^x|$; it shows the context-aware word features in $t^x$. By the two visualized features, we can know which parts of the context $t^y$ are more important for a word in sentence $t^x$, and a max-pooling, over those context-driven word representations, selects and forwards dominant \emph{(word, left$_\mathrm{context}$, right$_\mathrm{context}$, att$_\mathrm{context}$) combinations}  to the final decision maker.

Figure \ref{fig:visualization} shows the features\footnote{For simplicity, we show 2 out of 300 \modelname\ filters.} of sentence $t^x$ = ``\emph{A dog jumping for a Frisbee in the snow.}'' conditioned on the context $t^y$ = ``\emph{An animal is outside in the cold weather, playing with a plastic toy.}''. Observations: (i) The right figure shows that the attention mechanism successfully aligns some cross-sentence phrases that are informative to the textual entailment problem, such as ``dog'' to ``animal'' (i.e., $\mathbf{c}^x_{dog}\approx$ ``animal''), ``Frisbee'' to ``plastic toy'' and ``playing'' (i.e., $\mathbf{c}^x_{Frisbee}\approx$ ``plastic toy''+``playing''); (ii) The left figure shows a max-pooling over the generated features of filter\_1 and filter\_2 will focus on the  context-aware phrases (A, dog, jumping, $\mathbf{c}^x_{dog}$) and (a, Frisbee, in, $\mathbf{c}^x_{Frisbee}$) respectively; the two phrases are crucial to the entailment reasoning for this ($t^y$, $t^x$) pair.

\figref{visualizationpooling} shows the phrase-level (i.e.,
each consecutive tri-gram) attentions after the convolution
operation. As \figref{attentivepooling} shows, a subsequent pooling step will weight and sum up
those phrase-level hidden states as an overall sentence
representation. So, even though some phrases such as ``in
the snow'' in $t^x$ and ``in the cold'' in $t^y$ show
importance in this pair instance, the final sentence
representation still (i) lacks a fine-grained
phrase-to-phrase reasoning, and (ii) underestimates some
indicative phrases such as ``A dog'' in $t^x$ and ``An
animal'' in $t^y$.

Briefly, attentive convolution first
performs phrase-to-phrase, inter-sentence reasoning, then
composes features; attentive pooling composes phrase
features as sentence representations, then performs
reasoning. Intuitively, attentive convolution better fits
the way humans  conduct entailment reasoning, and
our experiments validate its superiority -- it is the
hidden states of the aligned phrases rather than their
matching scores that support better  representation
learning and decision making.

The comparisons in both \scitail\enspace and SNLI show that:

\textbullet\enspace CNNs with attentive convolution (i.e., \modelname) outperform the CNNs with attentive pooling (i.e., ABCNN and APCNN);

\textbullet\enspace Some competitors  got over-tuned on SNLI while
  demonstrating mediocre performance in \scitail\enspace --
  a real-world
  NLP task. Our system \modelname\enspace shows its robustness in both benchmark datasets.

\subsection{Sentence modeling with multiple contexts: claim verification}

\paragraph{Dataset.} For this task, we use FEVER
\cite{DBLPfever05355}; it infers the truthfulness of claims by extracted evidence.  The claims in FEVER were manually constructed
from  the introductory sections of about 50K popular Wikipedia articles in the
June 2017 dump. Claims have   9.4 tokens
on average. Table \ref{tab:feverdata} lists the claim
statistics.

\begin{figure}
  \setlength{\belowcaptionskip}{-5pt}
  \setlength{\abovecaptionskip}{0pt}
\centering
\includegraphics[width=0.45\textwidth]{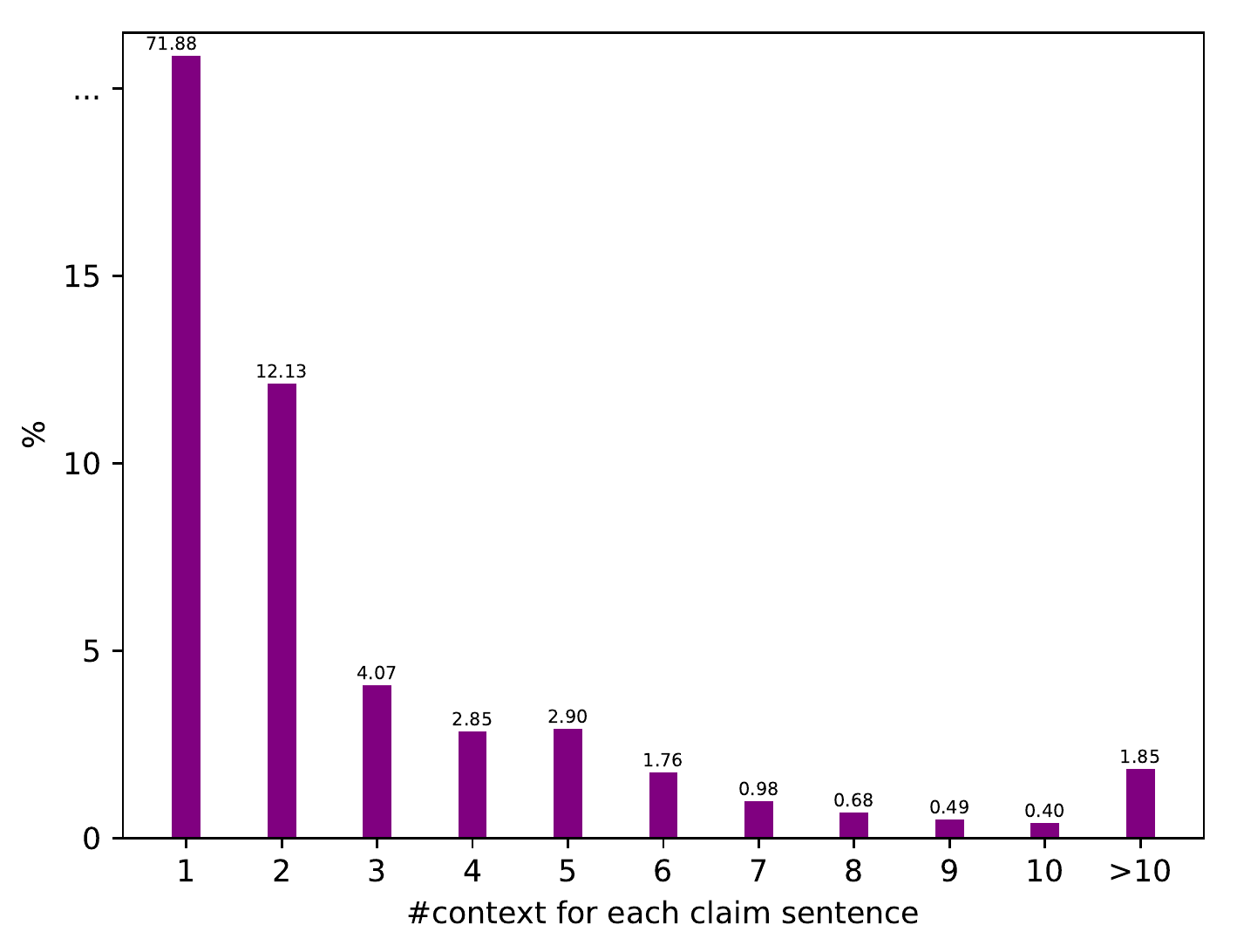}
\caption{Distribution of \#sentence  in FEVER evidence}\label{fig:evidistr}
\end{figure}

In addition to claims, FEVER also provides a
Wikipedia corpus  of approximately 5.4 million articles, from which gold evidences are gathered and provided. Figure \ref{fig:evidistr} shows the
distributions of sentence sizes  in FEVER's ground truth
evidence set (i.e., the context size in our experimental setup). We can see that roughly 28\% of evidence
instances cover more than one sentence and roughly 16\%
cover more than two sentences.


\begin{table}
  \centering
  \begin{tabular}{l|rrr}
   &  \#\textsc{Supported}  & \#\textsc{Refuted}  & \#NEI\\\hline
	train & 80,035 & 29,775 & 35,639\\
    dev & 3,333 & 3,333 & 3,333\\
    test & 3,333 & 3,333 & 3,333

\end{tabular}
\caption{Statistics of claims in FEVER dataset}\label{tab:feverdata}
\end{table}

Each claim is labeled as \textsc{Supported},
\textsc{Refuted} or \textsc{NotEnoughInfo} (NEI) given the gold evidence. The standard FEVER task also explores the performance of evidence extraction, evaluated by $F_1$ between extracted evidence and gold evidence. This work focuses on the claim entailment part, assuming the evidences are provided (extracted or gold). More specific, we treat a claim as $t^x$, and its evidence sentences as context $t^y$.

This task has two evaluations: (i) \textsc{All} -- accuracy of claim verification regardless of  the validness of evidence; (ii) \textsc{Subset} -- verification accuracy of a subset of claims, in which the gold evidence for \textsc{Supported} and \textsc{Refuted} claims must be fully retrieved.  We use the official evaluation toolkit.\footnote{\url{https://github.com/sheffieldnlp/fever-scorer}}

\paragraph{Setups.} (i) We adopt the same retrieved evidence
set (i.e, contexts $t^y$) as \newcite{DBLPfever05355}: top-5
most relevant sentences from top-5 retrieved wiki pages by a
document retriever \cite{DBLPChenFWB17}. The quality of this
evidence set against the ground truth is: 44.22 (recall),
10.44 (precision), 16.89 ($F_1$) on dev, and 45.89 (recall),
10.79 (precision), 17.47 ($F_1$) on test. This setup challenges
our system with potentially unrelated or even misleading
context. (ii) We use the ground truth evidence as
context. This lets us determine how far our \modelname\enspace
can go for this  claim verification problem once the
accurate evidence is given.

\begin{table}
\setlength{\tabcolsep}{3pt}
  \centering
  \begin{tabular}{lll|cc|c}
 & & & \multicolumn{2}{c|}{retrie. evi.} & gold\\
 & \multicolumn{2}{c|}{system} &  \textsc{All} & \textsc{Sub}  &  evi.\\\hline\hline
\multirow{13}{*}{\rotatebox{90}{\begin{tabular}{c}dev\end{tabular}}} &  \multicolumn{2}{l|}{MLP} & 41.86 & 19.04 & 65.13   \\
&\multicolumn{2}{l|}{Bi-CNN} & 47.82 & 26.99 & 75.02   \\
&\multicolumn{2}{l|}{APCNN} & 50.75 & 30.24 & 78.91   \\
&\multicolumn{2}{l|}{ABCNN} & 51.39 & 32.44 & 77.13   \\
&\multicolumn{2}{l|}{Attentive-LSTM} & 52.47 & 33.19 & 78.44   \\
 & \multicolumn{2}{l|}{Decomp-Att} & 52.09 & 32.57 & 80.82 \\
   & \multicolumn{5}{l}{\modelname}\\

  & \multicolumn{2}{l|}{\enspace\enspace light,context-wise} & 57.78 & 34.29 &  83.20\\
    & \multicolumn{2}{l|}{\enspace\enspace\enspace\enspace\enspace\enspace w/o conv.} & 47.29 & 25.94 &  73.18\\
  & \multicolumn{2}{l|}{\enspace\enspace light,context-conc} & 59.31 & 37.75 &  84.74 \\
    & \multicolumn{2}{l|}{\enspace\enspace\enspace\enspace\enspace\enspace w/o conv.} & 48.02 & 26.67 &  73.44\\
  & \multicolumn{2}{l|}{\enspace\enspace advan.,context-wise} & 60.20 & 37.94 &  84.99\\
  & \multicolumn{2}{l|}{\enspace\enspace advan.,context-conc} & \textbf{62.26} & \textbf{39.44} &  \textbf{86.02}\\
 \cline{1-6}

\multirow{2}{*}{\rotatebox{90}{\begin{tabular}{c}test\end{tabular}}} &  \multicolumn{2}{l|}{\cite{DBLPfever05355}} & 50.91 & 31.87 & --\\
& \multicolumn{2}{l|}{\modelname} &\textbf{61.03}& \textbf{38.77} & \textbf{84.61}
\end{tabular}
\caption{Performance on $dev$ and $test$ of FEVER. In ``gold
  evi.'' scenario, \textsc{All}  \textsc{Subset} are the same.}\label{tab:feverresults}
\end{table}
\paragraph{Baselines.} We first include the two systems explored by \newcite{DBLPfever05355}: (i) MLP: A multi-layer perceptron baseline with a single  hidden layer, based on tf-idf cosine similarity between the claim and the evidence \cite{DBLPRiedelASR17}; (ii) Decomp-Att
\cite{DBLParikhT0U16}: A decomposable attention model  that is tested in \scitail\enspace and SNLI before. Note that both baselines first relied on an IR system to extract the  top-5 relevant sentences from the retrieved top-5 wiki pages as \emph{ evidence} for claims, then  concatenated all evidence sentences  as a longer  context for a claim.

We then consider two variants of our \modelname\enspace in
dealing with modeling of $t^x$ with variable-size context $t^y$. (i) \emph{Context-wise}: we first use all evidence sentences one by one as context $t^y$ to guide the representation learning of the claim $t^x$, generating a group of context-aware representation vectors for the claim, then we do element-wise max-pooling over this vector group as the final representation of the claim. (ii) \emph{Context-conc}: concatenate all evidence sentences as a single piece of context, then model the claim based on this context. This is the same preprocessing step as \newcite{DBLPfever05355} did.
\begin{figure}
 \setlength{\belowcaptionskip}{-12pt}
 \setlength{\abovecaptionskip}{5pt}
\centering
\includegraphics[width=0.45\textwidth]{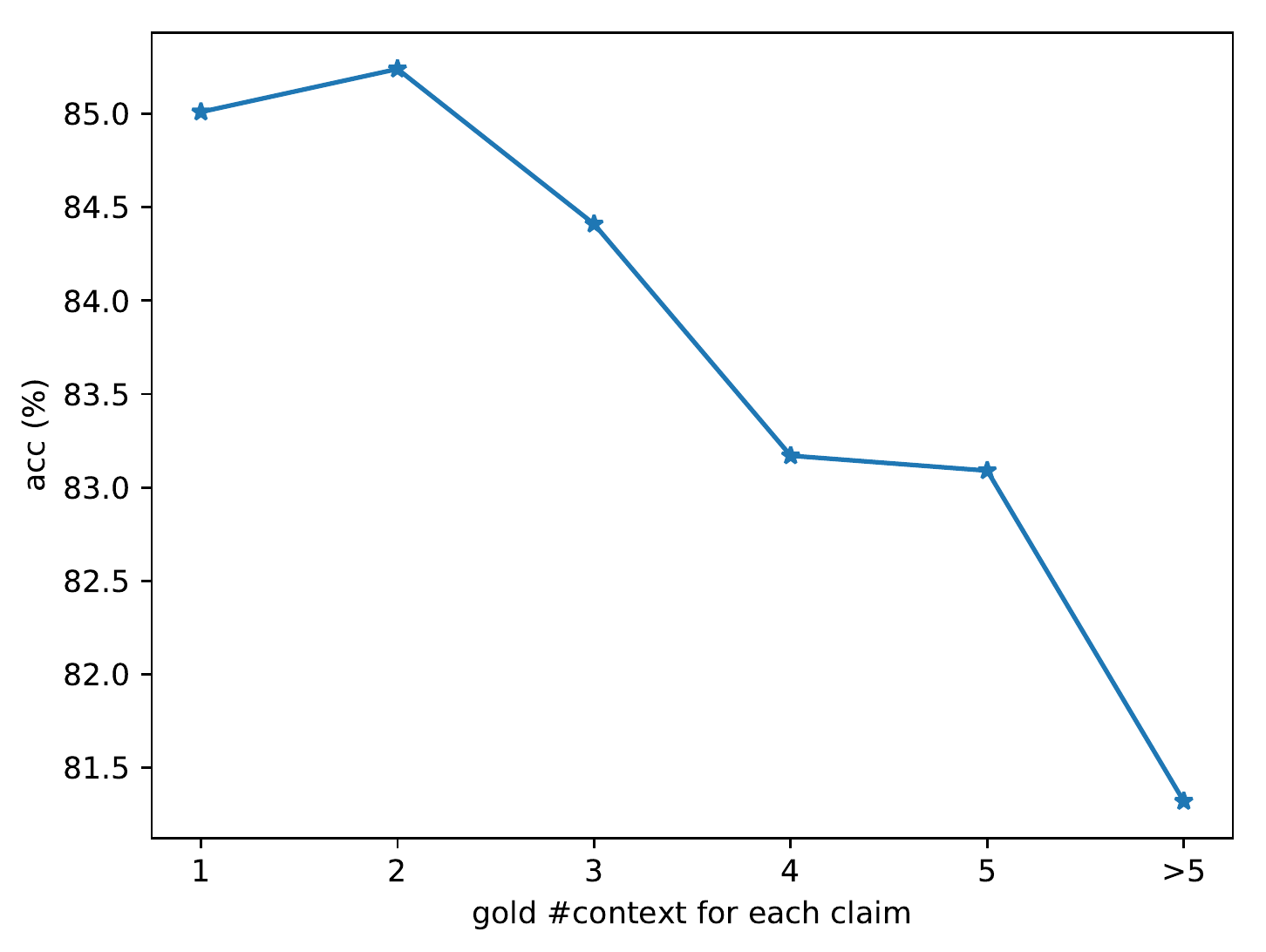}
\caption{Fine-grained \modelname\enspace performance  given variable-size golden FEVER evidence as claim's context}\label{fig:feverfine}
\end{figure}

\paragraph{Results.} Table \ref{tab:feverresults} compares our \modelname\enspace in different setups against the baselines. First, \modelname\enspace surpasses the top competitor ``Decomp-Att'', reported in \cite{DBLPfever05355}, with big margins in \emph{dev} (\textsc{All}: 62.26 vs. 52.09) and \emph{test} (\textsc{All}: 61.03 vs. 50.91).
In addition, ``advanced-\modelname'' consistently  outperforms its ``light'' counterpart. Moreover, \modelname\enspace surpasses attentive pooling (i.e., ABCNN\&APCNN) and ``attentive-LSTM'' by $>$10\% in \textsc{All}, $>$6\% in \textsc{Sub} and $>$8\% in ``\emph{gold evi.}''.

Figure \ref{fig:feverfine} further explores the fine-grained
performance of \modelname\enspace for different sizes of
gold evidence (i.e., different sizes of context $t^y$). The
system shows comparable performances for sizes 1 and 2.
Even for
context sizes  bigger than 5,
it only drops by 5\%.

Above experiments on claim verification clearly show the effectiveness of  \modelname\enspace  in sentence modeling with variable-size context. This should be attributed to the attention mechanism in \modelname, which enables a word or a phrase in the claim $t^x$ to ``see'' and accumulate all related clues even if those clues are scattered across multiple contexts $t^y$.

\paragraph{Error Analysis.} We do error analysis for  ``retrieved evidence''  scenario.

Error case \#1 is due to the failure of fully retrieving all evidence. For example, a successful \emph{support} of the claim ``\emph{Weekly Idol has a host born in the year 1978.}'' requires the information composition from three evidence sentences, two from the wiki article ``Weekly Idol'', one from  ``Jeong Hyeong-don''. However, only one of them is retrieved in the top-5 candidates. Our system predicts \textsc{refuted}. This error is more common in instances for which no evidence is retrieved.

Error case \#2 is due to the insufficiency of representation
learning. Consider the wrong claim ``Corsica belongs to
Italy'' (i.e., in \textsc{Refuted} class). Even though good evidence is retrieved, the system is misled by  noise evidence: ``\emph{It is located \ldots\ west of the Italian Peninsula, with the nearest land mass being the Italian island \ldots}''.

Error case \#3 is due to the lack of  advanced data
preprocessing. For a human, it is very easy to ``refute''
the claim ``Telemundo is a English-language television
network'' by the evidence ``\emph{Telemundo is an American
  Spanish-language terrestrial television \ldots}'' (from
the ``Telemundo'' wikipage), by checking the keyphrases:
``Spanish-language''
vs. ``English-language''. Unfortunately, both tokens are
unknown words in our system, as a result, they do not have
informative  embeddings. A more careful data preprocessing is expected to help.

\section{Summary}\label{sec:conclusion}
We presented \modelname,
the first work that enables CNNs to acquire the
attention mechanism commonly employed in RNNs.
\modelname\enspace combines the strengths of CNNs with the
strengths of
the RNN attention mechanism. On the one hand,
it makes
broad and rich context available for prediction, either
context from
external inputs (extra-context)   or internal inputs (intra-context).
On the other hand, it can take full advantage of the
strengths of convolution: it is more order-sensitive than
attention in  RNNs and local-context information can be
powerfully and efficiently modeled through convolution filters.
Our experiments demonstrate the effectiveness and flexibility of \modelname\enspace while modeling sentences with variable-size context.

\section*{Acknowledgments}
We gratefully acknowledge funding for this work by the European Research Council (ERC \#740516).
We would like to thank the anonymous reviewers
for their helpful comments.

\bibliography{tacl2018}
\bibliographystyle{acl_natbib}
\end{document}